\pdfoutput=1

\documentclass[11pt]{article}

 \usepackage{EMNLP2022}

\usepackage{times}
\usepackage{latexsym}
\usepackage{graphicx}
\usepackage[export]{adjustbox}
\usepackage{multirow}
\usepackage{array}
\usepackage{amsmath,bm}
\usepackage{amsfonts}

\usepackage{tikz}
\newcommand*\circled[1]{\tikz[baseline=(char.base)]{
            \node[shape=circle,draw,inner sep=0.7pt] (char) {#1};}}

\newcolumntype{L}[1]{>{\raggedright\let\newline\\\arraybackslash\hspace{0pt}}m{#1}}
\newcolumntype{C}[1]{>{\centering\let\newline\\\arraybackslash\hspace{0pt}}m{#1}}
\newcolumntype{R}[1]{>{\raggedleft\let\newline\\\arraybackslash\hspace{0pt}}m{#1}}
\usepackage{booktabs}
\usepackage[T1]{fontenc}
\usepackage{xcolor}
\usepackage{soul}
\usepackage[normalem]{ulem}
\definecolor{DarkGreen}{rgb}{0.0, 0.4, 0}
\definecolor{DarkYellow}{rgb}{0.4, 0.2, 0.0}
\definecolor{DarkPurple}{rgb}{0.44, 0.16, 0.39}
\definecolor{DarkRed}{rgb}{0.6,0,0}
\definecolor{DarkBlue}{rgb}{0,0,0.6}
\colorlet{LightYellow}{white!80!yellow}
\colorlet{LightRed}{white!80!red}
\colorlet{LightPurple}{white!80!purple}
\colorlet{LightBlue}{white!80!blue}
\colorlet{LightGreen}{white!80!green}
\definecolor{Red}{rgb}{1,0,0}
\definecolor{colorAdd}{HTML}{FFC2BA}

\usepackage{xcolor}
\makeatletter
\newcommand\hll{%
  \bgroup
  \UL@protected\def\sout{\bgroup \ULdepth =-.8ex \ULset}%
  \markoverwith{\textcolor{colorAdd}{\rule[-.5ex]{.1pt}{2.5ex}}}%
  \ULon}
\makeatother

\usepackage[utf8]{inputenc}

\usepackage{microtype}
\usepackage[colorinlistoftodos]{todonotes}

\usepackage{inconsolata}
\usepackage{enumitem}
\usepackage{subcaption}
\usepackage[algoruled, vlined ]{algorithm2e}

\usepackage{times}
\usepackage{import}
\usepackage{latexsym}
\usepackage{amsmath}
\usepackage{bbm}
\usepackage{url}
\usepackage{hyperref}
\usepackage{tabularx}
\usepackage{enumitem}

\usepackage{mathtools}
\usepackage{multirow}
\usepackage{comment}
\usepackage{float}
\usepackage{amsfonts}

\usepackage{subcaption}
\usepackage{cleveref}
\DeclareMathOperator*{\avg}{avg} %
\DeclareMathOperator*{\argmax}{argmax} %

\usepackage{inconsolata}

\newcommand{\arxivedits}{\textsc{arXivEdits}}

\title{{\arxivedits}: Understanding the Human Revision Process in \\ Scientific Writing}

\author{Chao Jiang\textsuperscript{1}, Wei Xu\textsuperscript{1}, Samuel Stevens\textsuperscript{2}\thanks{\hspace{4pt} Work done as an undergraduate student.} \\
\textsuperscript{1} School of Interactive Computing, Georgia Institute of Technology\\
\textsuperscript{2} Department of Computer Science and Engineering, Ohio State University\\

  {\small \tt chaojiang@gatech.edu \quad  wei.xu@cc.gatech.edu \quad stevens.994@osu.edu} 

}

\begin{document}
\maketitle
\begin{abstract}

Scientific publications are the primary means to communicate research discoveries, where the writing quality is of crucial importance. However,  prior work studying the human editing process in this domain mainly focused on the abstract or introduction sections, resulting in an incomplete picture. In this work, we provide a complete computational framework for studying text revision in  scientific writing. We first introduce {\arxivedits}, a new annotated corpus of 751 full papers from arXiv with gold sentence alignment across their multiple versions of revision, as well as fine-grained span-level edits and their underlying intentions for 1,000 sentence pairs. It supports our data-driven analysis to unveil the common strategies practiced by  researchers for revising their papers. To scale up the analysis, we also develop automatic methods to  extract revision at document-, sentence-, and word-levels. A neural CRF sentence alignment model trained on our corpus achieves 93.8 F1, enabling the reliable matching of sentences between different versions.  We formulate the edit extraction task as a span alignment problem, and our proposed method extracts more fine-grained and explainable edits, compared to the commonly used \verb|diff|  algorithm. An intention classifier trained on our dataset achieves 78.9 F1 on the fine-grained intent classification task. Our data and system are released at \url{tiny.one/arxivedits}.

\end{abstract}

\section{Introduction}

Writing is essential for sharing scientific findings. Researchers devote a huge amount of effort to revising their papers by improving the writing quality or updating new discoveries. Valuable knowledge is encoded in this  revision process. Up to January 1st, 2022, arXiv (\url{https://arxiv.org/}),  an open access e-print service,  has archived over  1.9 million papers, among which more than 600k papers have multiple versions available. This provides an amazing data source for studying text revision in scientific writing. Specifically, revisions between different versions of papers contain valuable information about logical  and structural improvements at document-level, as well as stylistic and grammatical refinements at sentence- and word-levels. It also can support various natural language processing (NLP) applications, including writing quality assessment and  error correction \cite{louis-nenkova-2013-makes,xue-hwa-2014-redundancy, daudaravicius-etal-2016-report, bryant-etal-2019-bea}, text simplification and compression \cite{xu-etal-2015-problems, filippova-etal-2015-sentence}, style transfer \cite{xu-etal-2012-paraphrasing, krishna-etal-2020-reformulating}, hedge detection \cite{medlock-briscoe-2007-weakly}, and paraphrase generation \cite{yao-multipit}.

 \begin{figure*}[t!]
    \centering
    \includegraphics[width=\linewidth]{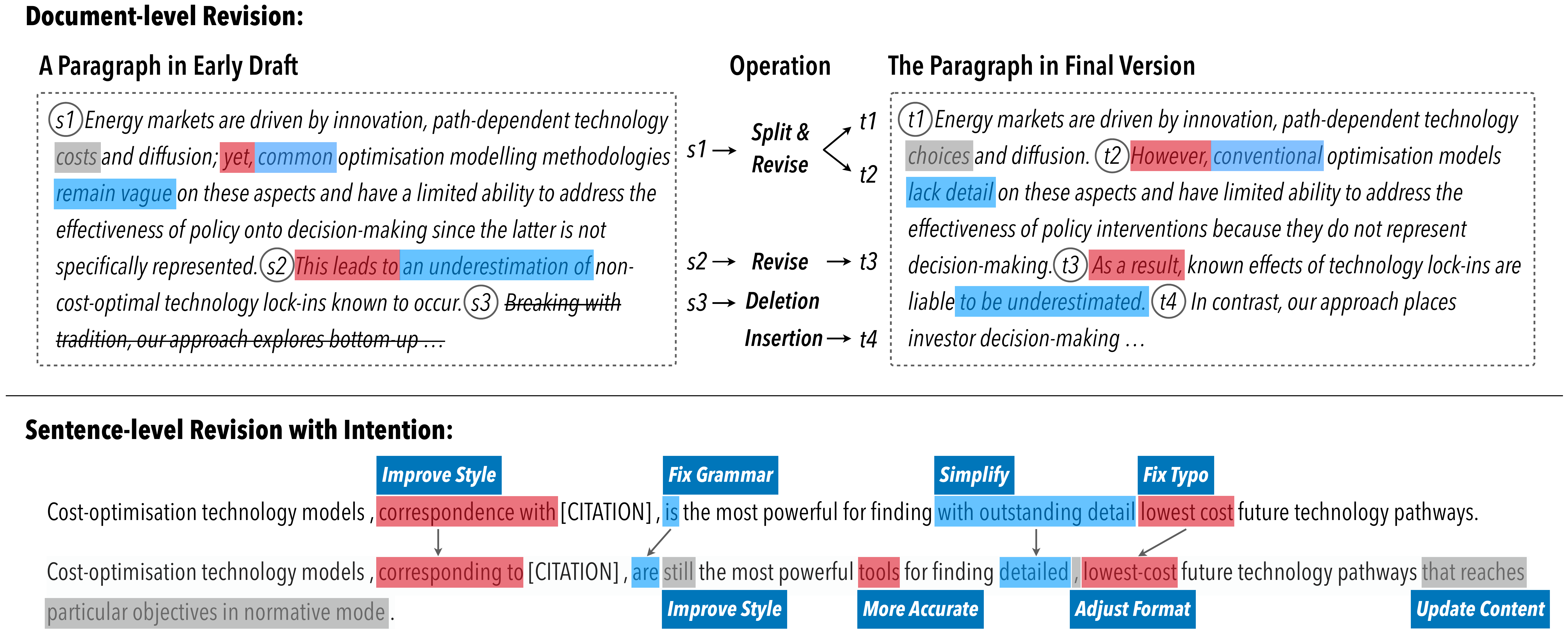}
    \vspace{-15pt}
    \caption{Our {\arxivedits} corpus consists of both document-level revision (top) and sentence-level revision with intention (bottom). The top part shows an aligned paragraph pair from the original and revised papers, where \circled{s1}  and \circled{t1} denote the corresponding sentences. For sentence-level revision, the fine-grained edits and each of their intentions are manually annotated.} 
    \label{fig:model_overview}
    \vspace{-15pt}

\end{figure*}

In this paper, we present a complete solution for studying the human revision process in the scientific writing domain, including annotated data, analysis, and system. We first construct {\arxivedits}, which consists of  751 full arXiv papers with gold sentence alignment across their multiple versions of revisions, as shown in Figure \ref{fig:model_overview}. Our corpus spans 6 research areas, including physics, mathematics, computer science, quantitative biology, quantitative finance, and statistics, published in 23 years (from 1996 to 2019). To the best of our knowledge, this is the first text revision corpus that covers full  multi-page research papers. To study sentence-level revision, we manually annotated fine-grained edits and their underlying intentions that reflect \textit{why} the edits are being made for 1,000 sentence pairs, based on a taxonomy that we developed  consisting of 7 categories.

Our dataset addresses two major limitations in prior work. First, previous researchers mainly focus on the abstract \cite{gabor-etal-2018-semeval,kang-etal-2018-dataset, du-etal-2022-understanding-iterative} and introduction \cite{tan-lee-2014-corpus, mita2022towards} sections, limiting the generalizability of their conclusions. In addition, a sentence-level revision may consist of multiple fine-grained edits made for different purposes (see an example in Figure \ref{fig:model_overview}). Whereas previous work either concentrates   on the change of a single word or phrase \cite{faruqui-etal-2018-wikiatomicedits, pryzant2020automatically} or extracts edits using the \verb|diff| algorithm \cite{myers1986ano}, which is based on minimizing the edit distance regardless of semantic meaning. As a result, the extracted edits are coarse-grained, and the intentions annotated on top of them can be ambiguous.

Enabled by our high-quality annotated corpus, we perform a series of data-driven studies to answer: \textit{what common strategies are used by authors to improve the writing of their papers?}  We also provide a pipeline system with 3 modules to automatically extract and analyze revisions at all levels. (1) A neural sentence alignment model trained on our data achieves 93.8 F1. It can be reliably used to extract parallel corpus for text-to-text generation tasks. (2) Within a revised sentence pair, the  edit  extraction is formulated as a span alignment task, and our method can extract more fine-grained and explainable edits compared to the \texttt{diff} algorithm. (3) An intention classifier trained on our corpus achieves 78.9 F1 on the fine-grained classification task, enabling us to scale up the analysis by  automatically extracting and classifying span-level edits from the unlabeled revision data.  We hope our work will inspire other researchers to further study the task of text revision in academic writing.

\section{Constructing {\arxivedits} Corpus}

In this section, we present the detailed procedure for constructing the  {\arxivedits}  corpus.
After posting preprints on arXiv, researchers can continually update the submission, and that constitutes the revisions.  More specifically, a revision denotes  two adjacent versions of the same paper.\footnote{For example,  the paper titled ``Attention Is All You Need'' (\url{https://arxiv.org/abs/1706.03762}) has  five versions on arXiv submitted by the authors, constituting four revisions (v1-v2, v2-v3, v3-v4, v4-v5).} An article group refers to all versions of a paper on arXiv (e.g., v1, v2, v3, v4). 
In this work, we refer to the changes applied to tokens or phrases within one sentence as sentence-level revision. The document-level revision refers to the change of an entire or several sentences, and the changes to the paragraphs can be derived from sentences.
Table \ref{table:arxiv_table} presents the statistics of document-level revision in our corpus.
After constructing this manually annotated corpus, we use it to train the 3 modules in our automatic system as detailed at $\$$\ref{sec:method}.

\subsection{Data Collection and Preprocessing}

We first collect metadata for all 1.6 million papers posted on arXiv between March 1996 and December 2019. We then randomly select 1,000 article groups from the 600k papers that have more than one versions available. To extract plain text from the LaTeX source code of these papers, we improved the open-source  OpenDetex\footnote{\url{https://github.com/pkubowicz/opendetex}} package to better handle macros, user-defined commands, and additional LaTeX files imported by the \textit{input} commands in the main file.\footnote{Our code is released at \url{https://tiny.one/arxivedits}} We find this method is less error-prone for extracting plain text, compared to using other libraries such as Pandoc\footnote{\url{https://pandoc.org/}}  used in \cite{cohan-etal-2018-discourse, roush-balaji-2020-debatesum}. Among the randomly selected 1,000 article groups, we obtained plain texts for  751  complete groups, with a total of 1,790 versions of papers, that came with the original LaTex source code and contained text content that was understandable without an overwhelming number of math equations. A breakdown of the filtered  groups is provided in Appendix \ref{sec:preprocessing}.

\subsection{Paragraph and Sentence Alignment}

 Sentence alignment can capture all document-level revision operations, including the insertion, deletion, rephrasing, splitting, merging, and reordering of sentences and paragraphs  (see Figure \ref{fig:model_overview} for an example).  Therefore, we propose the following 2-step annotation method to manually align sentences for papers in the 1,039 adjacent version pairs (e.g., v0-v1, v1-v2) from the 751 selected article groups, and the alignments between non-adjacent version pairs (e.g., v0-v2) then can be derived automatically.

\begin{enumerate}[topsep=2pt,itemsep=0pt,partopsep=2pt,parsep=1pt]
\item Align paragraphs using a light-weighted alignment algorithm that we designed based on Jaccard similarity \cite{jaccard1912distribution} (more details  in Appendix \ref{Appendix:para_align}). It can cover 92.1\% of non-identical aligned sentence pairs, based on a pilot study on 18 article pairs. Aligning paragraphs  first significantly reduces the number of sentence pairs that need to be annotated.

\item Collect annotation of sentence alignment  for every possible pair of sentences in the aligned paragraphs using Figure-Eight\footnote{\url{https://www.figure-eight.com/}}, a crowdsourcing platform. We ask 5 annotators to classify each pair into one of the following categories: \textit{aligned}, \textit{partially-aligned}, or \textit{not-aligned}.  Annotators are required to spend at least 25 seconds on each question. The annotation instructions and interface can be found in Appendix \ref{appendix:human-annotation}.  We embed one hidden test question in every five questions, and the workers need to maintain an accuracy over 85\% on the test questions to continue working on the task.
\end{enumerate}

We skip aligning  5.8\% sentences  that contain too few words or too many special tokens. They are still retained  in the dataset  for completeness, and are marked with a special token. More details about the annotation process  are  in Appendix \ref{sec:preprocessing} and \ref{Appendix:para_align}.  In total, we spent \$3,776 to annotate 13,008 sentence pairs from 751 article groups, with a 526/75/150 split for train/dev/test sets in the experiments of automatic sentence alignment  in 
$\$$\ref{sec:method}.  The inter-annotator agreement is 0.614 measured by Cohen's kappa \cite{artstein_inter-coder_2008}. To verify the crowd-sourcing annotation  quality, an in-house annotator manually aligns sentences for  10 randomly sampled groups  with 14 article pairs.  If assuming the in-house annotation is gold, the majority vote of crowd-sourcing annotation achieves an F1 of 94.2 on these 10 paper groups.

\subsection{Fine-grained Edits with Varied Intentions}

\renewcommand{\arraystretch}{1.1}
 \begin{table}[t!]
\centering
\small

\begin{tabular}{L{4.3cm}|C{2cm}}
\toprule

\textbf{Operation at Document-level} & Count  \\ [-0.7pt]
\midrule
\# of sent. insertion (0-to-1)  &  25,229        \\
\# of sent. deletion (1-to-0)  &  17,315       \\
\# of sent. rephrasing (1-to-1)  &    17,755    \\
\# of sent. splitting (1-to-n) &    378        \\
\# of sent. merging (n-to-1)  &   269       \\
\# of sent. fusion (m-to-n) & 142  \\ 
\# of sent. copying (1-to-1)  & 95,110   \\

\bottomrule

\end{tabular}
\vspace{-5pt}
\caption{Statistics of document-level revision in our {\arxivedits} corpus, based on manually annotated sentence alignment.}
\vspace{-15pt}
\label{table:arxiv_table}
\end{table}

Sentence-level revision involves the insertion, deletion, substitution, and reordering of words and phrases. Multiple edits may be tangled together in one sentence, while each edit is made for different purposes (see an example in Figure \ref{fig:model_overview}). Correctly detecting and classifying these edits is a challenging problem.  We first introduce the formal definition of edits and our proposed intention taxonomy, followed by the annotation procedure.

 \begin{table*}[pht!]
\renewcommand{\arraystretch}{1.2}
\small

\centering
\resizebox{\linewidth}{!}{
\begin{tabular}{@{\hspace{0.02cm}}L{3.8cm} @{\hspace{-0.25cm}} L{6.2cm}  @{\hspace{0.25cm}}  L{6.7cm} @{\hspace{0.10cm}}  R{0.9cm} @{\hspace{0.02cm}}}

\toprule
\textbf{Intention Label} & \textbf{Definition} & \textbf{Example} & \textbf{\%} \\
\toprule
\multicolumn{3}{@{\hspace{0.02cm}}l}{\textbf{Improve Language}} & 28.6\% \\
\midrule 
\quad More Accurate/Specific & Minor adjustment to improve the accuracy or specificness of the description. & Further, we suggest a relativistic-invariant protocol for quantum  \hll{\sout{information processing} communication}. & 11.5\%\\
\midrule 
\quad Improve Style & Make the text sound more  professional or coherent without altering the meaning. &  \dots due to hydrodynamic interactions among cells \hll{\sout{in addition with} besides} self-generated force \dots & 8.7\%\\
\midrule 
\quad Simplify &  Simplify complex concepts  or delete redundant  content to improve readability. & These include new transceiver architecture ( \hll{\sout{TXRU} array connected}  architecture ) \dots  & 7.6\%\\

\midrule 
\quad Other &  Other language improvements that don’t fall into the above categories. &  \dots due to changes in fuels used \hll{\sout{, or , in other words ,} associated} to changes of technologies . & 0.8\%\\

\midrule 
\textbf{Correct Grammar/Typo} & Fix grammatical errors, correct typos, or smooth out grammar needed by other changes. &  \hll{\sout{Not} Note} that the investigator might reconstruct each function \dots & 25.4\% \\

\midrule 
\textbf{Update Content} & Update large amount of scientific content, add or delete major fact. &  \dots characterized by  long range hydrodynamic term \hll{and self-generated force due to actin remodeling}.  & 28.8\% \\

\midrule 
\textbf{Adjust Format} & Adjust table, figure, equation, reference, citation, and punctuation etc. &  Similarly to what we did in \hll{\sout{Figure} Fig.} [REF] , the statistical results obtained by means of  \dots  & 17.2\% \\

    \bottomrule
    
\end{tabular}

}
\vspace{-5pt}
\caption{\label{tab:taxonomy}A taxonomy ($\mathcal{I}$) of edit intentions in scientific writing revisions. In each example, text with red background denotes the edit. Span with strike-through means the content got deleted, otherwise is inserted.}
\vspace{-15pt}
\end{table*}

\paragraph{Definition of Span-level Edits.} A sentence-level revision  $\mathcal{R}$ consists of the original sentence \(\bm{s}\), target  sentence \(\bm{t}\), and a series of fine-grained edits  $\bm{e}_i$. Each edit $\bm{e}_i$ is defined as a tuple $(\bm{s}_{a:b}, \bm{t}_{c:d}, \mathbf{I})$, indicating span $[s_a, s_{a+1},..., s_{b}]$ in the original sentence is transformed into span $[t_c, t_{c+1},..., t_{d}]$ in the  target sentence, with an intention label $\mathbf{I} \in \mathcal{I}$ (defined in Table \ref{tab:taxonomy}). The type of edit can be recognized  by spans $\bm{s}_{a:b}$ and $\bm{t}_{c:d}$, where  $\bm{s}_{a:b}=\mathrm{[NULL]}$ indicating insertion, $\bm{t}_{c:d}=\mathrm{[NULL]}$  for deletion, $\bm{s}_{a:b}=\bm{t}_{c:d}$ representing reordering, and $\bm{s}_{a:b} \neq \bm{t}_{c:d}$ for substitution.

\paragraph{Edit Intention Taxonomy.}

We propose a new taxonomy to comprehensively capture the intention of text revision in the scientific writing domain, as shown in Table \ref{tab:taxonomy}. Each edit is classified into one of the following categories: \textit{Improve Language}, \textit{Correct Grammar/Typo}, \textit{Update Content}, and \textit{Adjust Format}.  Since our goal is to improve the writing quality, we further break the \textit{Improve Language} type into  four fine-grained categories.  During the design, we extensively consult prior literature in text revision \cite{faigley1981analyzing, fitzgerald1987research,daxenberger2016writing}, edit categorization \cite{bronner-monz-2012-user,yang-etal-2017-identifying-semantic}, and analysis in related areas such as Wikipedia \cite{daxenberger-gurevych-2013-automatically} and argumentative essays \cite{zhang-etal-2017-corpus}. The taxonomy is improved for several rounds based on the feedback from four NLP researchers and two in-house annotators with linguistic background.

\paragraph{Annotating Edits.} 

 In  pilot study, we found that directly annotating fine-grained edits is a tedious and complicated task for annotators, as it requires separating and matching edited spans across two sentences.  To assist the annotators, we use monolingual word alignment \cite{lan-etal-2021-neural},  which can find the correspondences between words and phrases with a similar meaning in two sentences, as an intermediate step  to reduce the cognitive load during annotation. We find that, compared to strict word-to-word matching, edits usually have larger granularity and may cross linguistic boundaries. For example, in Figure \ref{fig:model_overview}, ``corresponding to'' and ``correspondence with'' should be treated as a whole to  be meaningful and  labeled an intention. Therefore, the edits can be annotated by adjusting the boundaries of the span alignment. We propose the following 2-step method that leverages word alignment to assist the annotation of edits:

\begin{enumerate}[topsep=2pt,itemsep=0pt,partopsep=2pt,parsep=1pt]

\item Collect word alignment annotation by asking in-house annotators to manually correct the automatic word alignment generated by the neural semi-CRF word alignment model \cite{lan-etal-2021-neural}. The aligner is trained on the MTRef dataset and achieves state-of-the-art performance on the monolingual word alignment task with 92.4 F1. 

\item Annotate edits by having in-house annotators inspect and correct the fine-grained edits that are extracted from word alignment using simple heuristics. The heuristics are detailed in $\$$\ref{sec:edit_extraction_experiment}. Two principles are followed during the correction: (1) Each edit should  have a clear intention and relatively clear phrase boundaries; (2) Span pairs in substitution should be semantically related, otherwise should be treated as separated insertion and deletion.

\end{enumerate}

 We manually annotate insertion, deletion, substitution, and derive  reordering automatically, since it can be reliably found by heuristics. Due to the slight variance in granularity, it is possible that more than one answer is acceptable. 
 Therefore, we include all  the alternative edits  for sentence pairs in the dev and test sets in our annotation, among which 16\% have more than one answer. 
 
 Overall, we found that our method can annotate  more accurate and fine-grained edits compared to prior work that uses the \texttt{diff} algorithm. The \texttt{diff} method is based on minimizing the edit distance regardless of  semantic meaning. Therefore, the extracted edits are coarse-grained and may contain many errors  (detailed in Table \ref{table:identification_variations}).

 \begin{figure}[hpt!]
\centering
\vspace{-5pt}
\begin{subfigure}{.46\linewidth}
  \centering
  \includegraphics[width=\linewidth]{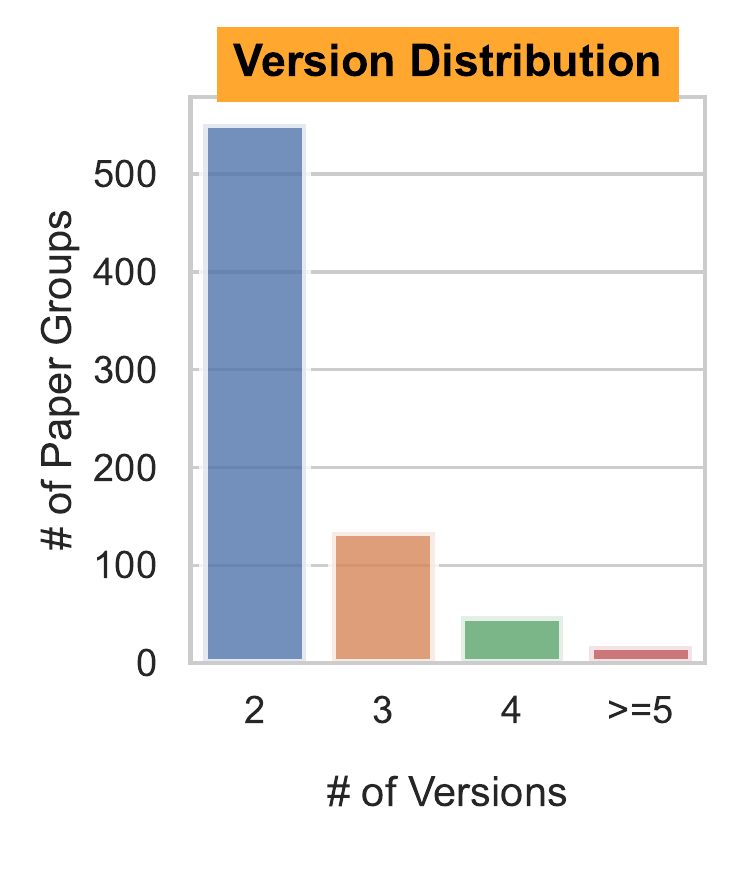}
  \label{fig:sub1}
\end{subfigure}%
\begin{subfigure}{.54\linewidth}
  \centering
  \includegraphics[width=\linewidth]{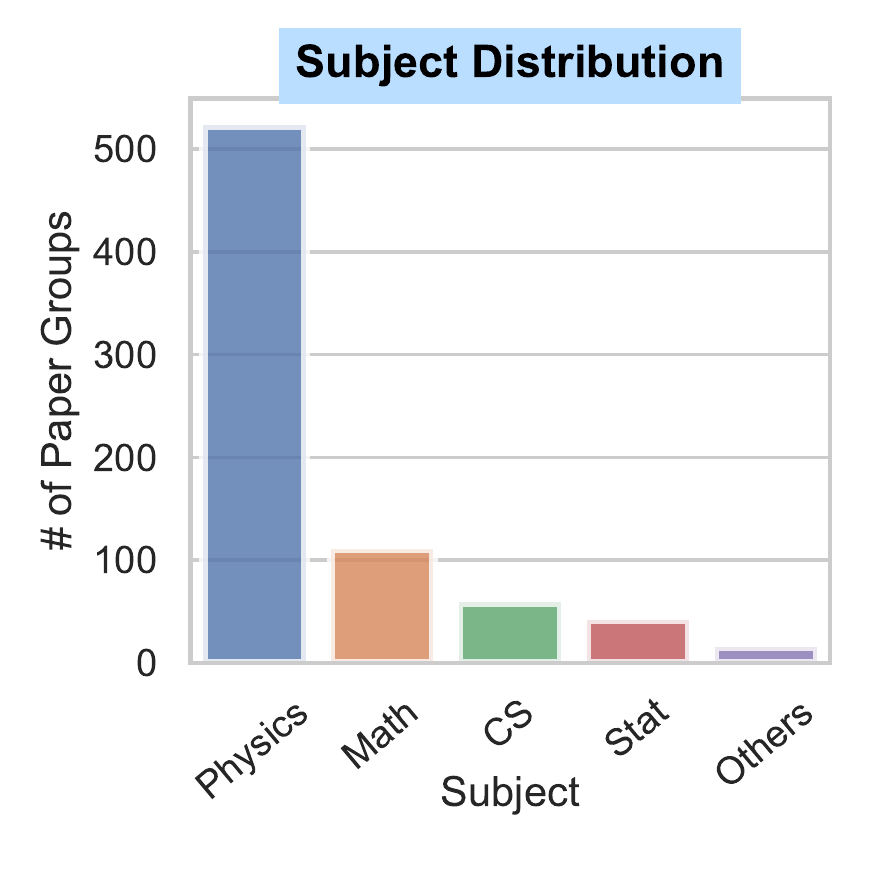}
  \label{fig:sub2}
\end{subfigure}
\vspace{-25pt}
\caption{Distribution of versions (left) and subjects (right) for papers in our corpus.}
\vspace{-7pt}
\label{fig:distribution}
\end{figure}

\vspace{-5pt}

\begin{figure}[bht!]
    \centering

\includegraphics[width=0.93\linewidth]{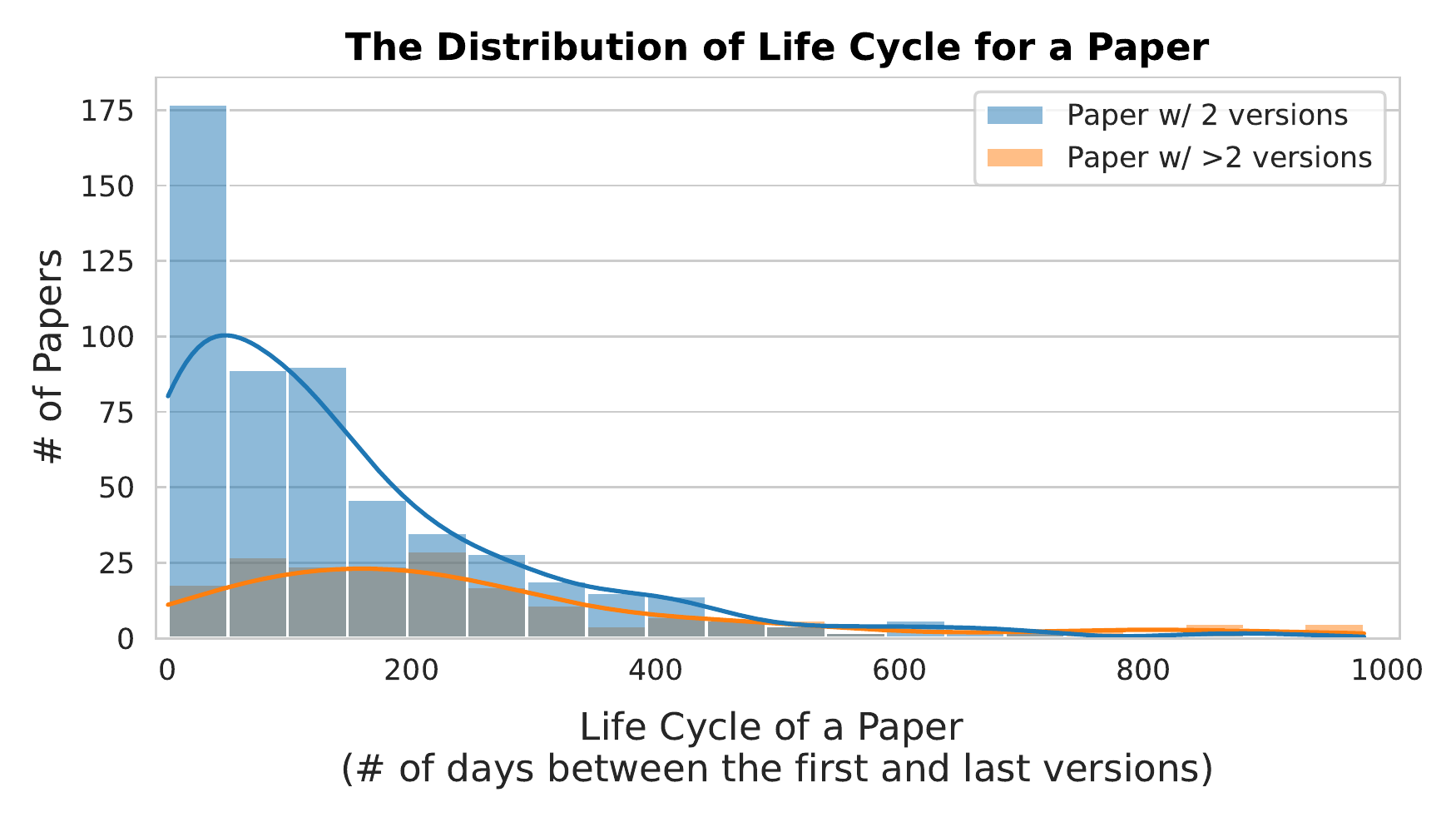}

\vspace{-10pt}

    \caption{The life cycle of each paper, measured by the time interval between the first and the last versions.} 
    \label{fig:life_cycle}
    
\vspace{-10pt}
\end{figure}

\paragraph{Annotating Intention.} As intentions can differ subtly, correctly identifying them is a challenging task.  Therefore, instead of crowdsourcing, we hire two experienced in-house annotators to annotate the intention for 2,122 edits in 1,000 sentence revisions. A two-hour training session is provided to both annotators, during which they are asked to annotate 100 sentence pairs and discuss until consensus. The inter-annotator agreement is 0.67 measured by Cohen Kappa \cite{artstein_inter-coder_2008}, and 0.81 if collapsing the \textit{Improve Language}  category.  The 1,000 sentence pairs are split into 600/200/200 for train/dev/test sets in experiments.

\section{Analysis of Document-level Revisions}

As a distinct style, scientific writing needs to be  clear, succinct, precise, and logical. To understand \textit{what common strategies are used by authors to improve the writing of their papers},  we present a data-driven study on document-level revisions in the scientific writing domain. This is enabled by our high-quality manually annotated corpus that consists of 1,790 versions of  751 full  papers across 6 research areas in 23 years.

\subsection{Distribution of Subjects and Versions}

\begin{figure}[bht!]
    \centering

\begin{subfigure}{0.5\textwidth}
\includegraphics[width=0.93\linewidth]{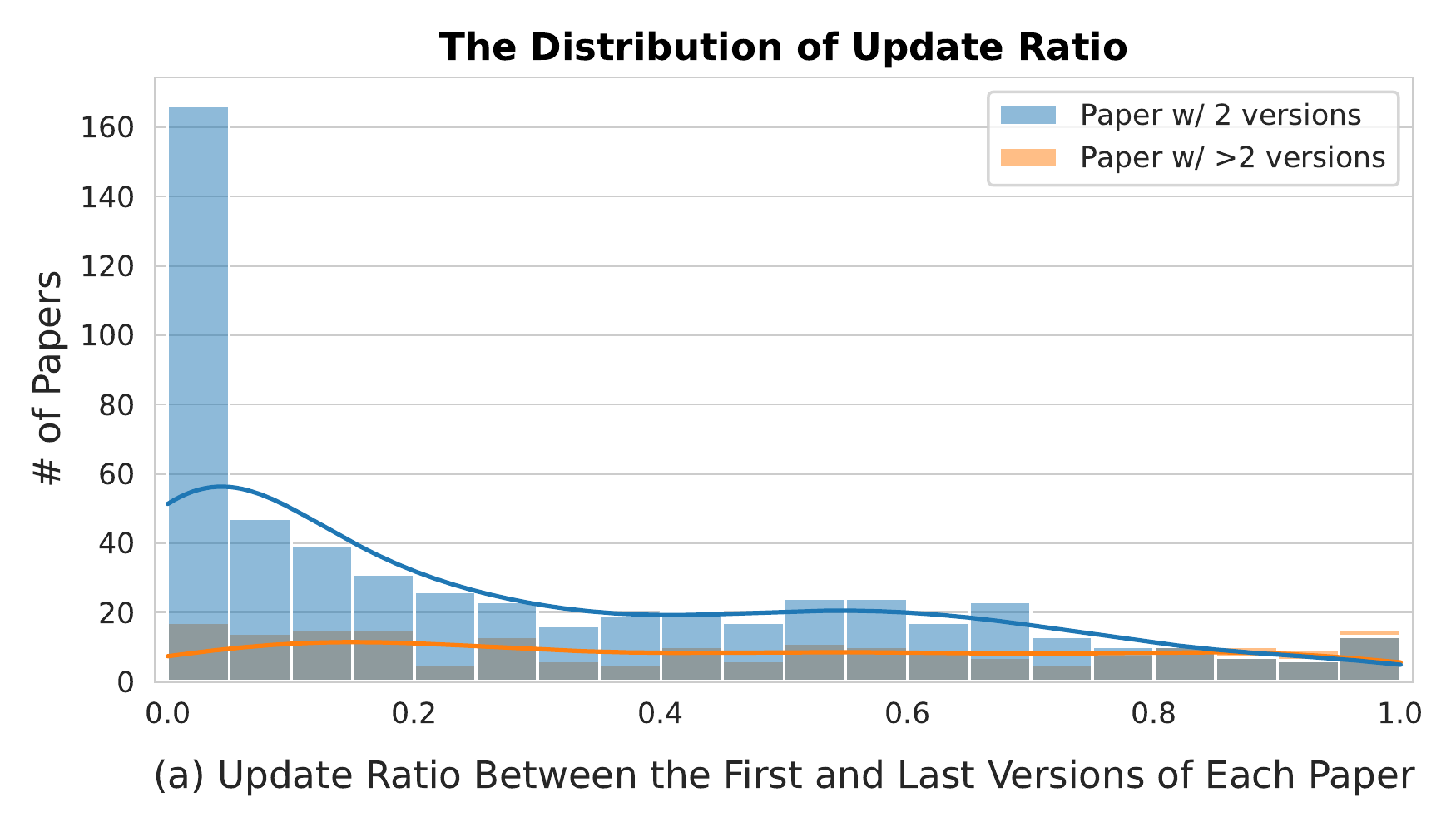}
\end{subfigure}

\begin{subfigure}{0.5\textwidth}
\includegraphics[width=0.93\linewidth]{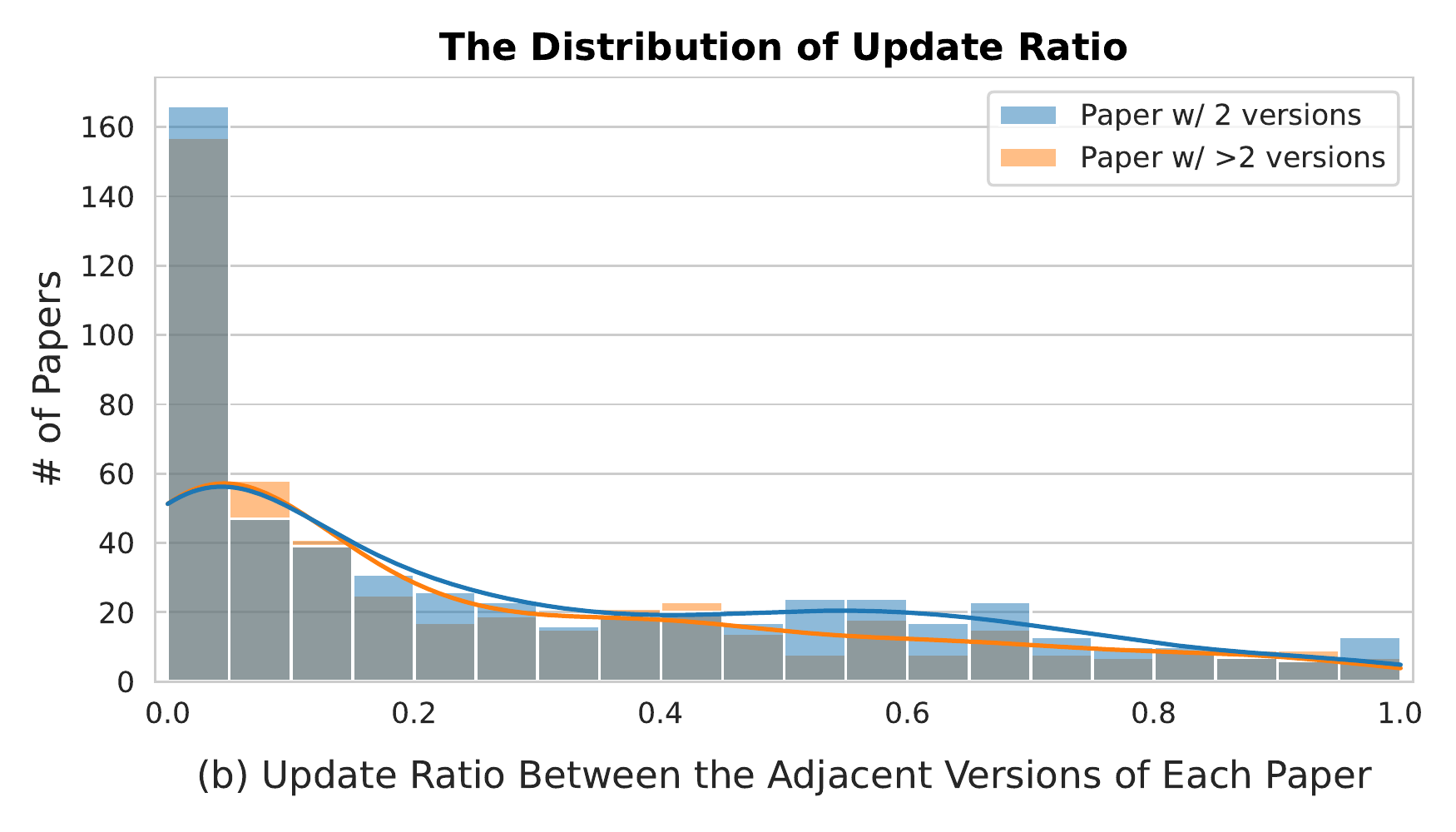}
\end{subfigure}

    \caption{The distribution of update ratio. The figure above demonstrates that papers with more versions are more likely to undergo a significant revision in their life cycle. While the two types of papers have a similar distribution of update ratio between adjacent versions, as shown in the figure below.} 
    \vspace{-15pt}
    \label{fig:update_ratio}

\end{figure}

Figure \ref{fig:distribution} plots the statistics for the  paper subjects and the number of versions. Physics (69.7\%) and Math (14.8\%) have the largest volume of  multi-version papers, mainly due to the long history of use and a large number of sub-fields.  About 26.7\%  papers have more than 2 versions available, enabling the study of iterative revisions. Figure \ref{fig:life_cycle} plots the length of the life cycle for each paper in our corpus,  demonstrating a long-tail distribution.

\subsection{Analysis of the Overall Update Ratio}

We first investigate, in general, how much content is being updated for each paper during its life cycle, which can potentially affect the type of revisions contained therein.   We define the \textit{Update Ratio} as 1 minus the percentage of sentences being kept between two versions, which is derived from manually annotated sentence alignment.

Figure \ref{fig:update_ratio}(a) presents how much content is being updated for each paper between its first and last versions. For papers that have two versions available, the distribution is  heavily skewed towards the left end. The median update ratio is 19.0\%, meaning that most papers have a  mild revision. Whereas the distribution is much flatter for papers with multiple versions, indicating they are more likely to have a major revision in the life cycle.  Interestingly, a peak appears at the tail of the distribution, which means 3.7\% of the papers are almost completely rewritten.  However, as shown in Figure \ref{fig:update_ratio}(b), both types of papers have a similar distribution of update ratio for revisions between adjacent versions.

\begin{figure}[pht!]
    \centering
    \includegraphics[width=0.8\linewidth]{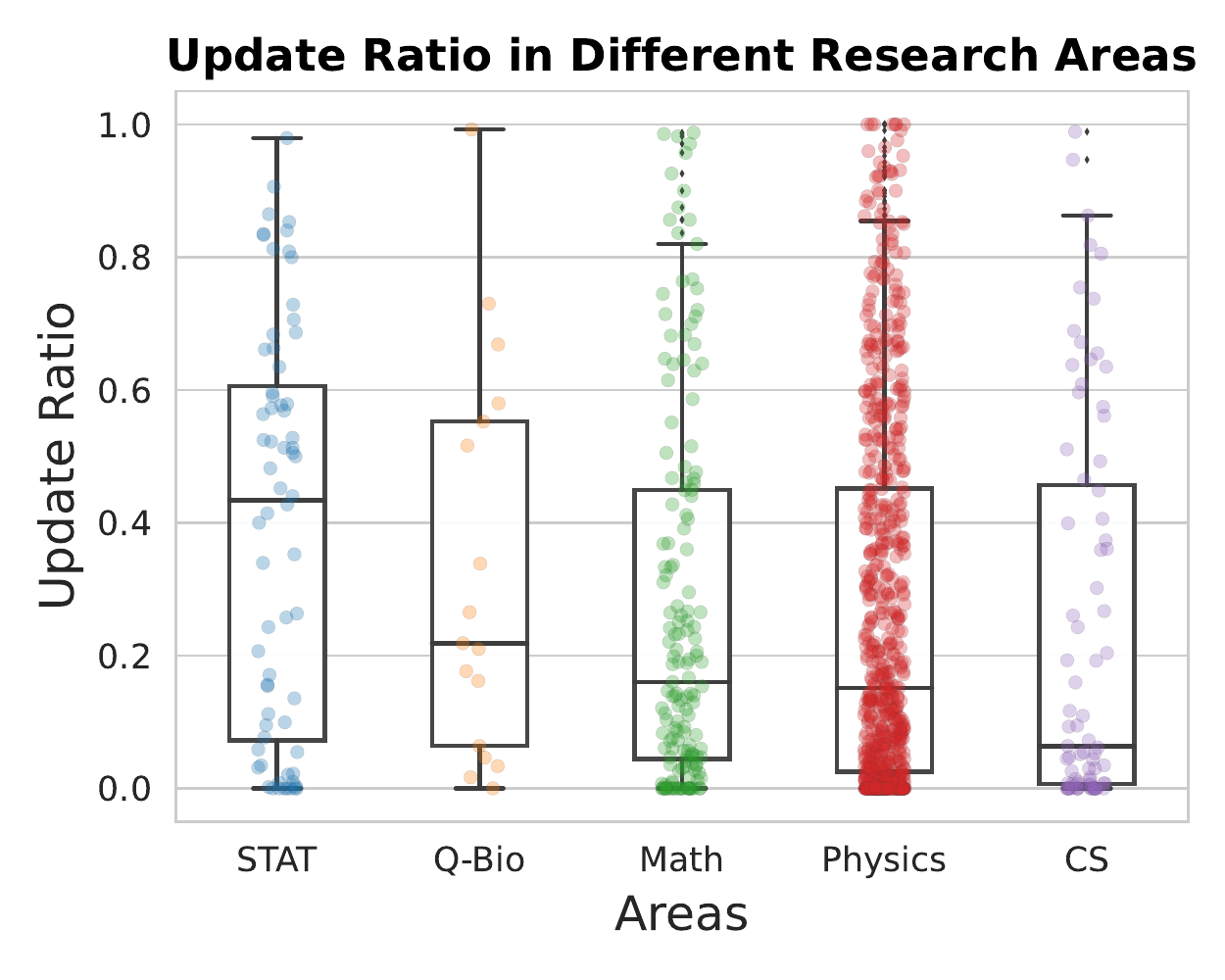}
    \vspace{-15pt}
    \caption{Update ratio for papers in different research areas. Papers in STAT have higher update ratios compared to papers in CS.} 
    \label{fig:sfs}
    \vspace{-10pt}
\end{figure}

 \begin{figure}[hpt!]
\centering
  
  \includegraphics[width=\linewidth]{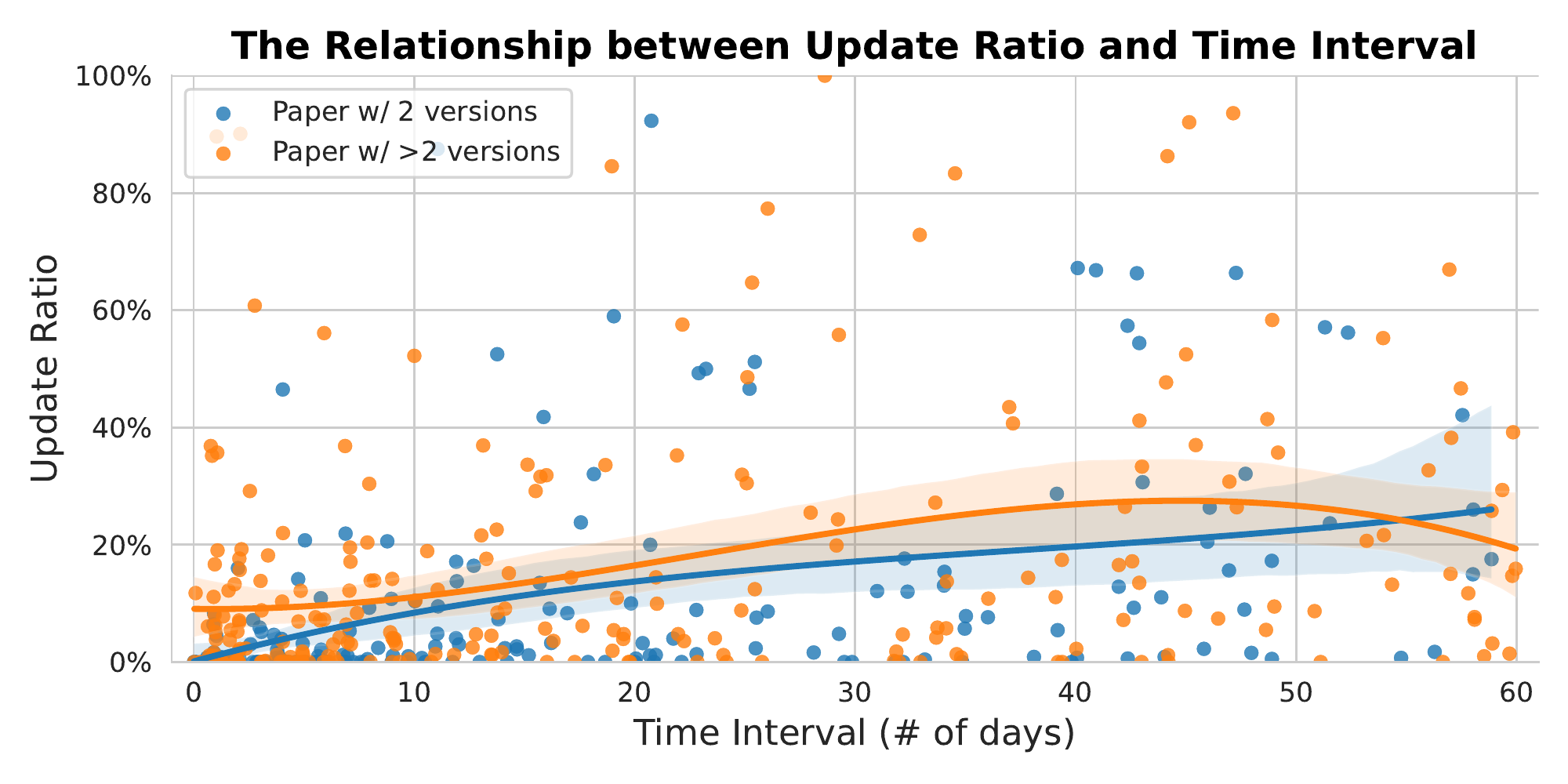}
\vspace{-25pt}
\caption{The relationship between update ratio  and time  between adjacent submissions.}
\label{fig:relationship}
\vspace{-14pt}

\end{figure}

\paragraph{Research Areas.} We hypothesize researchers in different areas may have different practices for revising their papers. Figure 4 visualizes the distribution of update ratio for papers on different subjects. Researchers in Statistics make more significant  revisions to their papers compared to the CS area.

\paragraph{Time Interval.} Intuitively, the time interval between submissions may correlate with the overall update ratio. We calculate the Pearson's correlation between the update ratio and the time spent on the revision, which is measured by the difference in timestamps between adjacent submissions.  The correlation values are 0.577 and   0.419  for  papers that have two versions and multiple versions available, and both correlations are significant. Figure \ref{fig:relationship} visualizes the relationship. Researchers make quick submissions for small adjustments while spending more time on major revisions.

\setlength{\tabcolsep}{3pt}

\subsection{Analysis of the Updated Sentences}
 \begin{figure}[hpt!]
\centering
  
  \includegraphics[width=\linewidth]{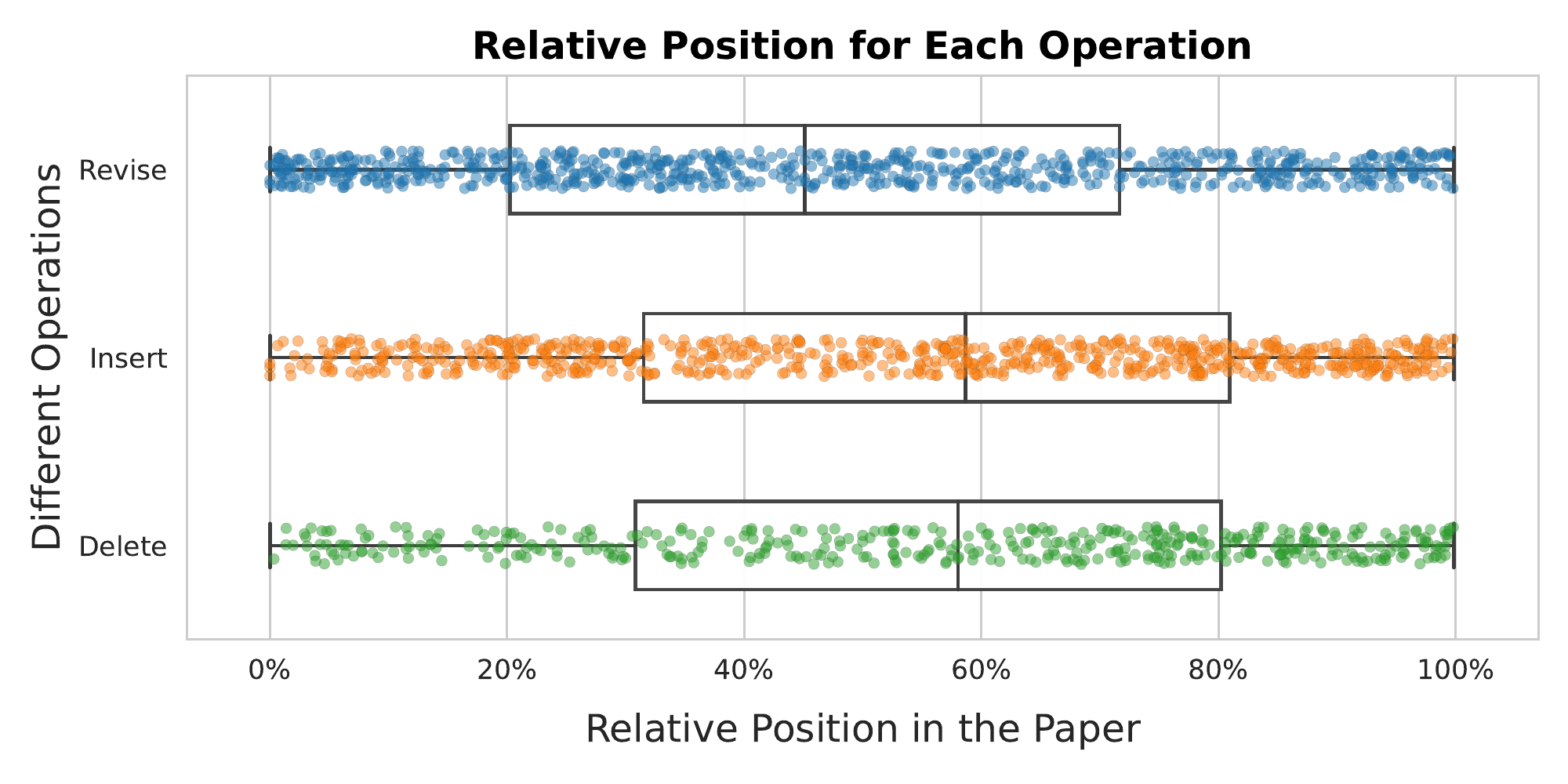}
\vspace{-25pt}
\caption{The relative position of the sentences that are being inserted, deleted, and revised.}
\label{fig:distribution_ins_del_sub_1}
\vspace{-10pt}

\end{figure}

 \begin{figure}[hpt!]
\centering
  
  \includegraphics[width=\linewidth]{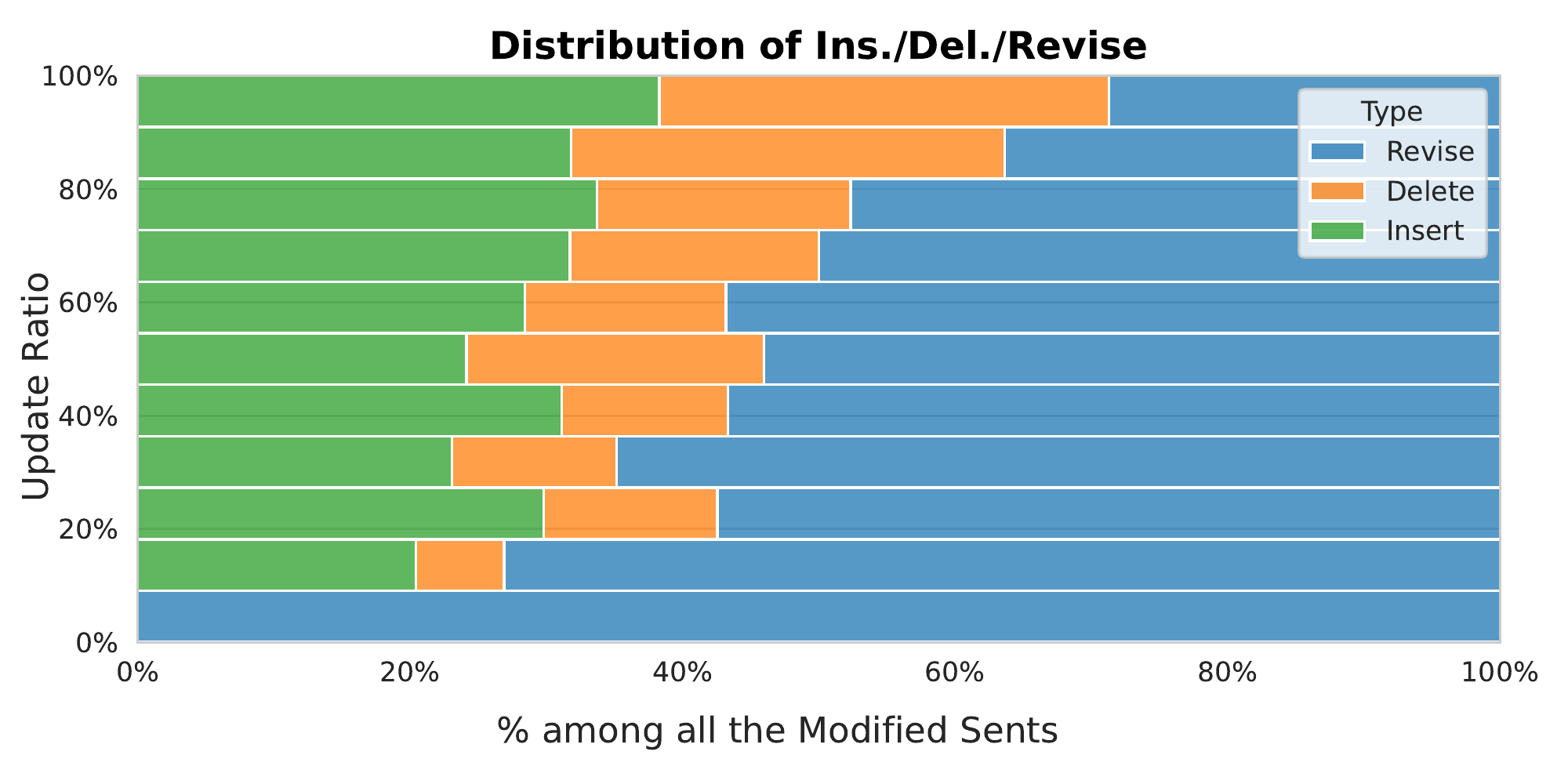}

\vspace{-10pt}
\caption{The composition of edit actions as the update ratio changes.}
\label{fig:distribution_ins_del_sub_2}
\vspace{-15pt}

\end{figure}

We  explore the dynamic of document-level edit operations to answer: \textit{where will and how  researchers update the sentences in their papers?}   The relative positions of the inserted, deleted, and revised sentences are visualized in Figure \ref{fig:distribution_ins_del_sub_1}. Researchers, in general, revise more sentences at the beginning of a paper, while the insertion and deletion of sentences occur more in the  latter parts. This makes sense because the abstract and introduction sections  are usually  frequently  revised by  the authors, since they are among the most important sections.  As shown in Figure \ref{fig:distribution_ins_del_sub_2}, revised sentences take the majority when  update ratio is low. As more content is being modified, the insertion and deletion of sentences  will become more dominant, which is likely to correspond to the  major updates  on the main body of  papers.

\subsection{Analysis of the Edit Intention}

\begin{figure}[pht!]
\vspace{-5pt}
    \centering
    \includegraphics[width=\linewidth]{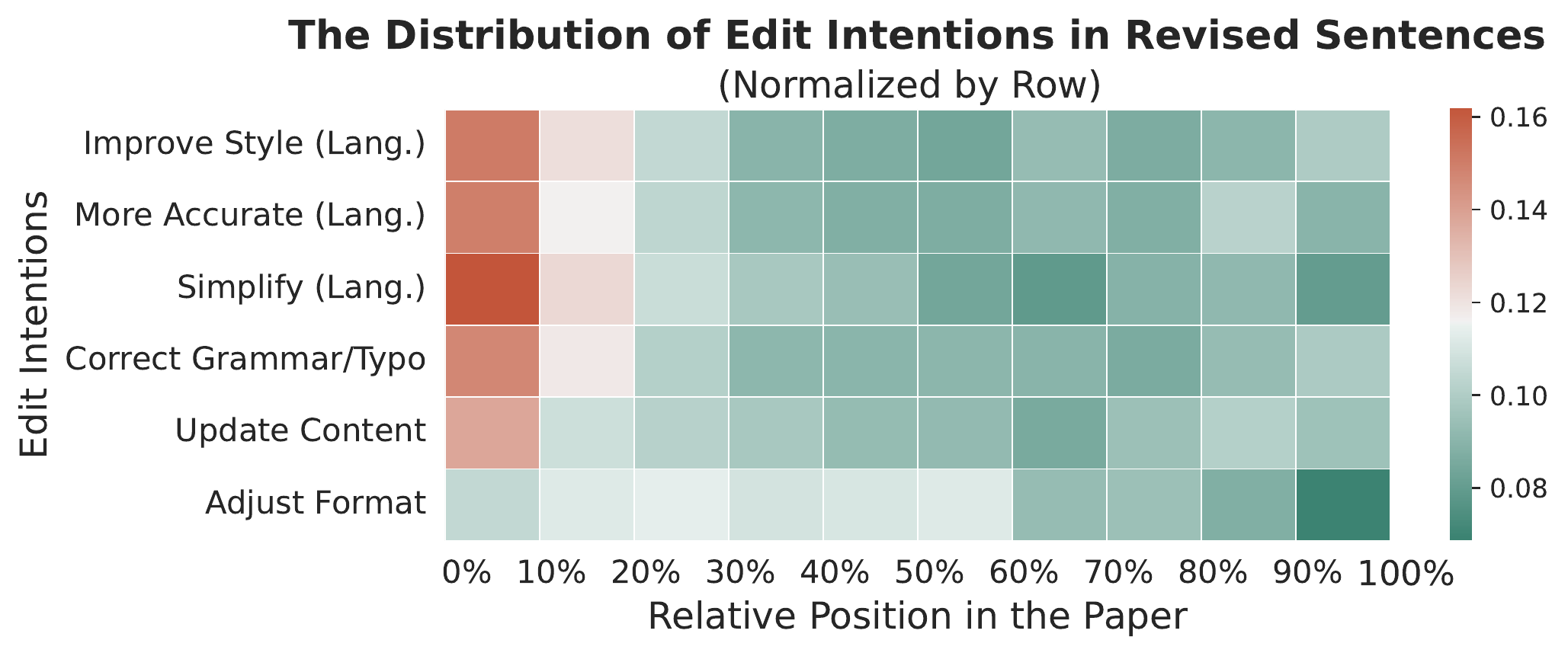}
    \vspace{-20pt}
    \caption{The distribution of intentions for span-level edits in the revised sentences in our corpus. } 

    \label{fig:intention_dis}
    \vspace{-15pt}
\end{figure}

To understand \textit{why} the researchers revised the sentences, we run our span-level edit extraction and intention classification system (details in \S \ref{sec:method}) on all the revised sentences between adjacent versions in 751 article groups. The distribution of the intentions  is visualized in Figure \ref{fig:intention_dis}. Most of the language-related edits occur at the beginning of a paper. The aggregation is gradually reduced for grammar/typo- and content-related edits. The adjustments to  format (punctuations, figures, tables, citations, etc.) span throughout the whole paper.

\begin{table*}[t]
\renewcommand{\arraystretch}{1.3}
\small
\centering
\begin{tabular}{l || cccc ||cccc | ccc  | ccc  }

\toprule
\multirow{2}{*}{\textbf{Methods}}   & \multicolumn{4}{c||}{\textbf{Perf. on  $\leq$ 5 edits}} &  \multicolumn{4}{c|}{\textbf{Perf. on All Revisions}} &  \multicolumn{3}{c|}{\textbf{\% of Edit Types}} &  \multicolumn{3}{c}{\textbf{Len. of Edits}}  \\

 &  P &  R &  F1 &  EM  &  P &  R &  F1 &  EM &  Ins. &  Del. & Sub. &  Ins. &  Del. & Sub.    \\

    \midrule

    Semi-CRF Aligner$_{simple}$ & 89.8 & 90.1 & 90.0 & 85.9 & 87.5 & 87.7 & 87.6& 80.5 & 32.9 & 26.7 & 40.4 & 4.66 & 4.98 & 2.21   \\

    Semi-CRF Aligner$_{parse}$ & 90.0 & 90.0 &90.0 & \textbf{87.0} & 87.4 & 86.8 & 87.1 & \underline{81.5} & 32.7 & 25.0 & 42.3 & 4.76 & 5.17 & 2.72 \\

    QA-align$_{simple}$ &  \underline{90.3} &  \textbf{90.9} & \textbf{90.6} & \textbf{87.0} & \underline{87.7} & \textbf{88.4} & \underline{88.0} &  \textbf{82.0}  & 33.2 & 24.0 & 42.9 & 4.46 & 4.62 & 2.08  \\
    QA-align$_{parse}$ & \textbf{90.4} & \underline{90.7} & \underline{90.5} & \underline{86.5} & \textbf{88.1} & \underline{88.1} & \textbf{88.1} & \underline{81.5} & 32.6 & 23.5 & 43.8 & 4.65 & 4.24 & 2.49  \\
    
     \midrule
    Latexdiff & 79.9 & 78.6 & 79.3 & 75.7 & 76.2 & 74.3 & 75.3 &  70.0 & 26.2 & 14.4 & 59.3  & 3.89  & 4.27 & 4.73   \\

    \bottomrule
\end{tabular}
\caption{Performance of different edit extraction methods on the {\arxivedits} testset. The \textbf{Len.} is measured by the number of tokens. We report performance on all sentence revisions, and on sentence pairs with $\leq$ 5 edits, which takes 92.5\% of the data. The best and second best scores in each column are highlighted by  \textbf{bold} and \underline{underline}.}
\vspace{-10pt}
\label{table:identification_variations}
\end{table*}

\section{Automatic Edit Extraction and Intention Identification} 
\label{sec:method}
As manual annotation is costly and time-consuming, we develop a pipeline system to automatically analyze the revisions at scale. Our system consists of sentence alignment, edit extraction, and intention classification modules, which are trained and evaluated on our annotated data. The methods  and  evaluation results of each step are detailed below. Example outputs from our system are presented in Figure \ref{fig:system_output}.

\subsection{Edits Extraction via Span Alignment}
\label{sec:edit_extraction_experiment}

Prior work relies on diff algorithm to extract edits, which is based on string matching regardless of semantic meaning. To extract more fine-grained and explainable edits, we formulate the edit extraction as a span alignment problem. Given the original and revised sentences, the fine-grained edits are derived from span alignment using simple heuristics.

\paragraph{Our Method.}
We finetune two state-of-the-art word alignment models: neural semi-CRF model \cite{lan-etal-2021-neural} and  QA-Aligner  \cite{nagata-etal-2020-supervised}   on our {\arxivedits} corpus, after  train them on the MTRef dataset \cite{lan-etal-2021-neural} first. Although sourced from the news domain, we find finetuning the models on MTRef, which is the largest monolingual word alignment corpus, helps to improve  4 points on the F1 score. When fine-tuning on   {\arxivedits}, the annotated edits are used as training labels, where substitutions  are formulated as span alignment, insertions and deletions are the unaligned tokens, and the rest words  will be aligned to their identical counterparts.

\begin{figure}[pht!]
    \centering
    \includegraphics[width=0.95\linewidth]{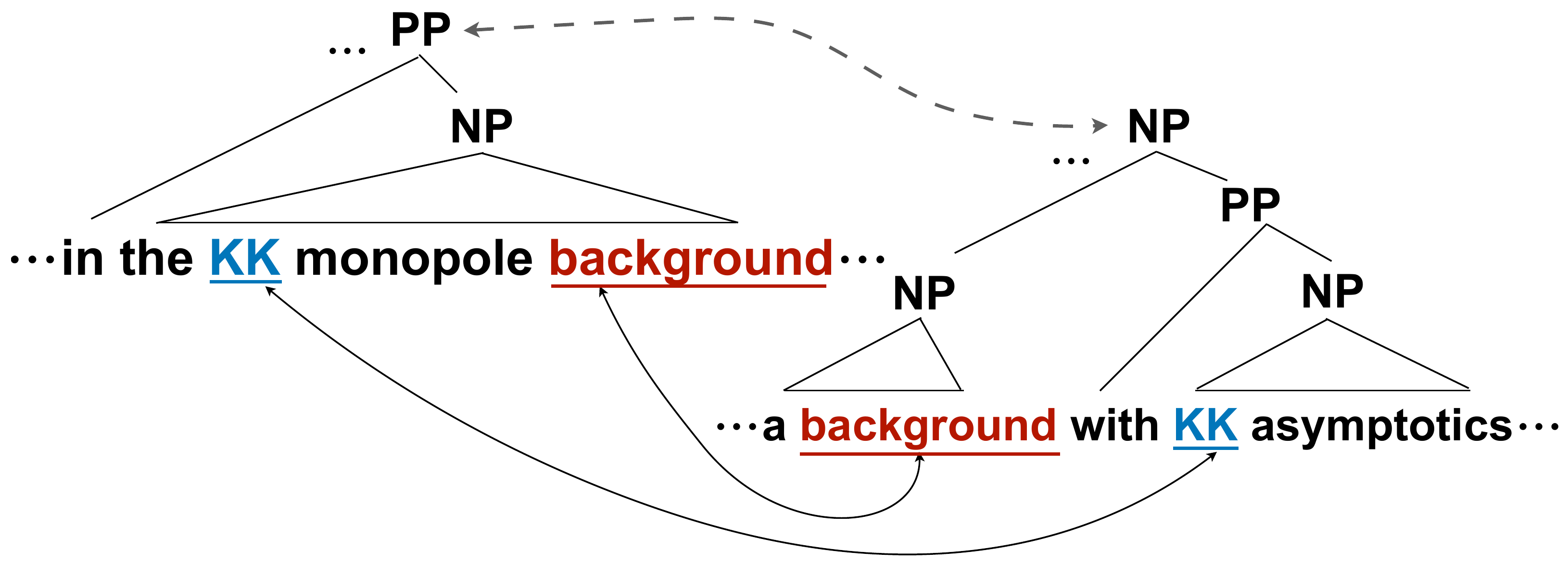}
    \caption{Illustration for extracting edits by leveraging constituency parsing tree. In this example, two full spans that only have loose correspondence can be aligned.} 
    \vspace{-15pt}
    \label{fig:tree_method}
\end{figure}

When running inference, the output edits are derived from span alignment using simple heuristics, where the insertions and deletions are unaligned  tokens in the revised and original sentences, respectively. Substitutions are the non-identical span alignments. A simple post-processing step is applied  to strip the identical words at the beginning and  end of the substituted span pairs.

To enable more flexible granularity, we also design slightly more complex heuristics to extract edits by leveraging compositional span alignment. As shown in  Figure \ref{fig:tree_method}, for each aligned word in two sentences, we iteratively traverse their parent nodes in two constituency parsing trees \cite{joshi-etal-2018-extending} for \textit{max-level} times to find the lowest ancestors in two trees that can resolve all the involved word alignment without conflict. Instead of separated word-to-word replacements, the two spans will be treated as a whole in the substitution. The \textit{max-level} is a hyperparameter and  can be adjusted to control the granularity of the extracted edits.

\paragraph{Baseline}  Diff \cite{myers1986ano} algorithm have been widely used in prior work to extract edits from text revision \cite{ yang-etal-2017-identifying-semantic, du-etal-2022-understanding-iterative}. It is an unsupervised method based on dynamic programming for finding the longest common subsequences between two strings.  Insertions and deletions are derived from unmatched tokens.  Substitutions are derived from adjacent insertion and deletion pairs. Diff algorithm has many implementations with different heuristics for post-processing. We compare against its implementation  in the latexdiff package, which is used in a recent work \cite{du-etal-2022-understanding-iterative}.

\begin{table}[phb!]
\vspace{-10pt}
\renewcommand{\arraystretch}{1.1}
\small
\centering
\begin{tabular}{l | cc  | cc }

\toprule
\multirow{2}{*}{\textbf{Models}} &  \multicolumn{2}{c|}{\textbf{4-Class}} &  \multicolumn{2}{c}{\textbf{8-Class}} \\

  & Accuracy & Weighted F1 & Accuracy & Weighted F1 \\

    \midrule
    \multicolumn{5}{l}{\textit{Trained w/ 8-class}}\\
    \midrule
    
    PURE & 69.8 & 69.6 & 66.5 & 65.4  \\ 
    T5-base & 74.2 & 73.5 & \underline{68.6} & \underline{66.4}   \\
    T5-large & \textbf{84.4} & \underline{84.4} & \textbf{79.3} & \textbf{78.9} \\
    \midrule
    \multicolumn{5}{l}{\textit{Trained w/ 4-class}}\\
    \midrule
    
    PURE & 72.1 & 72.0 & -- & -- \\ 
    T5-base & \underline{77.4} & 77.3 & -- & -- \\
    T5-large & \textbf{84.4} & \textbf{84.6} & -- & -- \\
    
    \bottomrule
\end{tabular}
\vspace{-5pt}
\caption{Performance of intention classification  on the {\arxivedits} testset.} 
\label{table:intention-inentification}
\end{table}

\paragraph{Results} We report precision, recall, F1, and exact match (EM) for edit extraction. Table 2 presents the results on  {\arxivedits} testset. We report the  performance on all 200 sentence pairs in the test set, and on a subset of sentence pairs with $\leq$ 5 edits, which take 92.5\% of the entire data and are more common in real applications. Using  simple heuristics,  both models finetuned on our dataset outperform the baseline method by more than 10 points in F1 and EM. In addition,    enabling compositional span alignment  by leveraging the constituency parsing tree can increase the granularity of the extracted edits, as shown in  the ``Len. of Edits'' column. For the latexdiff method, about 59.3\% of extracted edits are span substitutions, with an average length of 4.73 tokens. This is because the diff method derives edits by minimizing the edit distance. Combining with the post-processing heuristics, latexdiff treats  everything as large chunk substitutions regardless of their semantic similarity.

\subsection{Intention Classification}

Given an edit and  the original/revised sentences, the goal here is to classify its edit intention.  We formulate it in a way  that is similar to the relation extraction task. We experiment with two competitive models: T5 \cite{JMLR:v21:20-074} and PURE \cite{zhong-chen-2021-frustratingly}. The input is the concatenation of two sentences, where the edited spans are surrounded by special markers with the type (ins./del./subst.). The PURE model predicts the intention by classification, and the T5 model will generate the intention string.

\begin{table}[htb!]

\renewcommand{\arraystretch}{1.3}
\small
\centering
\begin{tabular}{L{3cm}C{1.3cm}C{1.1cm}C{1cm} }

\toprule
\textbf{Intention Label} & \textbf{Precision} & \textbf{Recall} & \textbf{F1}  \\
\midrule

Adjust Format & 96.7 & 94.6 & 95.6 \\
Update Content & 84.8 & 86.9 & 85.8  \\
Fix Grammar/Typo & 81.1 & 85.1  &83.1 \\
Language-Simplify & 75.0 & 66.7 & 70.6  \\
Language-Accurate & 54.7 & 63.0 & 58.6\\
Language-Style  & 46.9 & 37.5 & 41.7  \\

    \bottomrule
\end{tabular}
\vspace{-4pt}
\caption{Breakdown performance of the best performing T5-large model on  {\arxivedits} testset for fine-grained intention classification task. }
\vspace{-16pt}
\label{table:fine-grain-intention}
\end{table}

 \begin{figure*}[pht!]
    \centering
    \includegraphics[width=\linewidth]{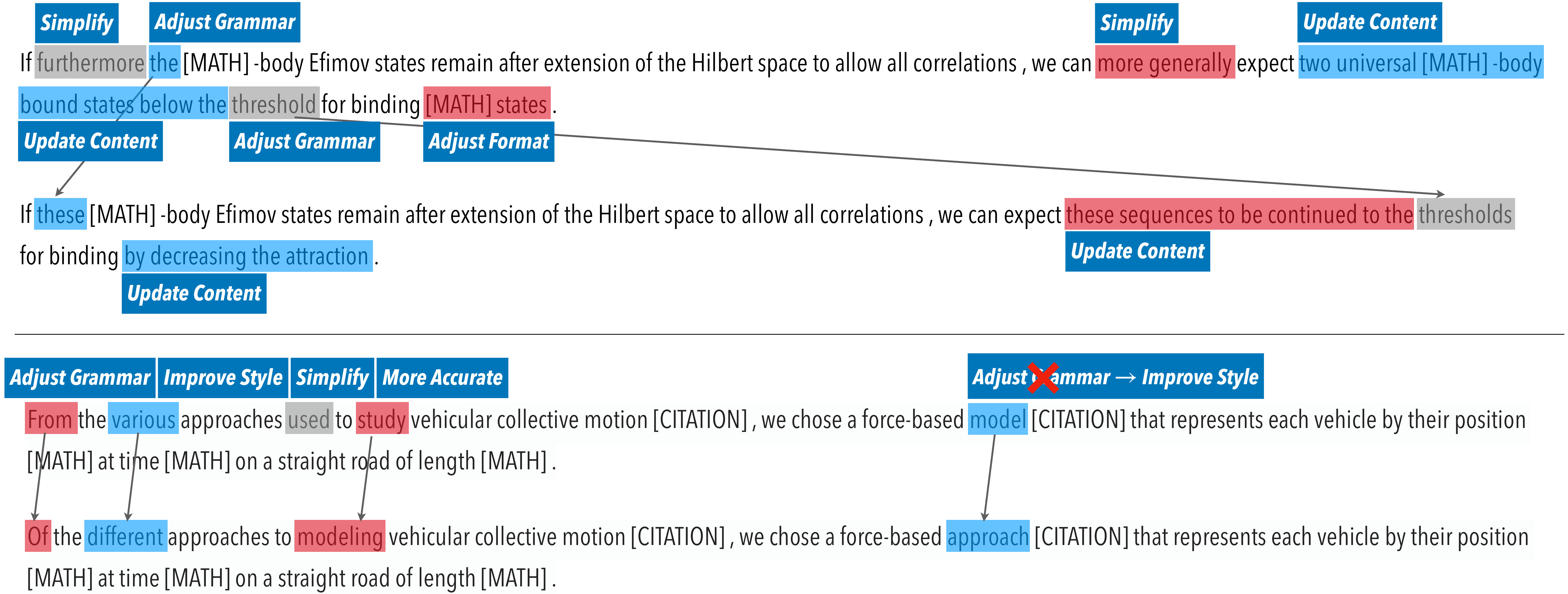}
    \vspace{-20pt}
    \caption{Two example outputs from our Semi-CRF Aligner$_{simple}$ system. The intentions are predicted by our best-performing T5 model.} 
    \label{fig:system_output}
    \vspace{-15pt}

\end{figure*}

\paragraph{Results.} Table \ref{table:intention-inentification} shows the results for both fine-grained and coarse-grained (collapsing the \textit{Improve Language} category)  classification experiments.  Collapsing labels helps to improve the  performance in the 4-way classification task, where a T5-large model achieves an accuracy of 84.4. Though it's challenging to pick up the differences between  7 types of intentions, the T5-large model trained with  fine-grained labels achieves an accuracy of 79.3. The per-category performance of the best-performing T5 model is presented in Table \ref{table:fine-grain-intention}.  It performs well in separating top-layer categories. Within the  \textit{Improve Language} type, it also achieves reasonable performance on \textit{Accurate} and \textit{Simplify} categories, while fall short on \textit{Style}, which is likely due to the inherited difficulty in identifying language style.

\subsection{Sentence Alignment}

Accurate sentence alignment is crucial for reliably tracking document-level revisions. Prior work mainly relies on surface-level similarity metrics, such as BLEU score \cite{faruqui-etal-2018-wikiatomicedits, faltings-etal-2021-text} or Jaccard  coefficients \cite{xu-etal-2015-problems}, combined with greedy or dynamic programming algorithms to match sentences. Instead, we finetune a supervised neural CRF alignment model on our annotated dataset. The neural CRF aligner is shown to achieve better performance at aligning sentences from articles with different readability levels in the Newsela Corpus \cite{jiang-etal-2020-neural}.

\paragraph{Our Methods.} We first align paragraphs using the  light-weighted paragraph alignment algorithm we designed (more details in  Appendix \ref{Appendix:para_align}). Then, for each aligned paragraph pair,  we apply our trained neural CRF alignment model to align sentences  from both the old  to  new version and the  reversed directions. The outputs from both directions are  merged  by intersection.

\paragraph{Results}

\setlength{\tabcolsep}{4pt}
\begin{table}[hpt!]
\small

\centering
\resizebox{\linewidth}{!}{%
\renewcommand{\arraystretch}{1.3}
\begin{tabular}{lccc}
\toprule
\textbf{Methods}  & \textbf{Precision} & \textbf{Recall} & \textbf{F1} \\
\toprule
Char. 3-gram (\citeauthor{stajner-etal-2018-cats})  & 87.7 & 87.7 & 87.7 \\
TF-IDF (\citeauthor{paetzold-etal-2017-massalign})  & 90.3 & \textbf{91.6} & \underline{90.9}  \\
Jaccard (\citeauthor{xu-etal-2015-problems}) & \underline{90.7} & 89.5 & 90.1  \\
BLEU (\citeauthor{faruqui-etal-2018-wikiatomicedits}) & 89.9 & 89.6 & 89.8 \\

\midrule
Neural CRF Aligner$_{Dual}$ (Ours) & \textbf{96.9} & \underline{91.0} & \textbf{93.8} \\

\bottomrule
\end{tabular}
}
\caption{\label{table:sentence_alignment_result}Evaluation Results of different sentence alignment methods on our {\arxivedits} testset.}
\vspace{-15pt}
\end{table}
\setlength{\tabcolsep}{3pt}

We report precision, recall, and F1 on the binary classification task of \textit{aligned + partially-aligned} vs. \textit{not-aligned}. Table \ref{table:sentence_alignment_result} presents the experimental results on {\arxivedits} testset. It is worth noticing that the identical sentence pairs are excluded during the evaluation as they are trivial to classify and will inflate the performance. For the similarity-based method, we tune a threshold based on the maximal F1 on the devset.  By training the state-of-the-art neural CRF sentence aligner on our dataset and merging the output from both directions, we are able to achieve 93.8 F1, outperforming other methods by a large margin. It is worth noticing that the precision of our model is particularly high, indicating that it can be reliably used to extract high-quality aligned sentence pairs, which can be used as the training corpus for downstream text-to-text generation tasks.

\section{Related Work}

\paragraph{Text Revision in Scientific Writing.} As a clear and concise style of writing, various aspects of scientific writing has been studied in previous work, including style \cite{bergsma-etal-2012-stylometric},  quality \cite{louis-nenkova-2013-makes}, hedge \cite{medlock-briscoe-2007-weakly}, paraphrase \cite{dong-etal-2021-parasci}, statement strength \cite{tan-lee-2014-corpus}, and grammar error correction \cite{daudaravicius-etal-2016-report}. Prior work studying scientific writing mainly focuses on the abstract and introduction sections \cite{tan-lee-2014-corpus, du-etal-2022-understanding-iterative, mita2022towards}. In comparison, we develop methods to annotate and automatically analyze  full research papers. Our work  mainly focuses on the writing quality aspect.

\paragraph{Edit and Edit Intention.} Previous work in studying the human editing process  \cite{faruqui-etal-2018-wikiatomicedits, pryzant2020automatically} mainly focuses on the change of a single word or phrase, as it is hard to pair complex edits in both sentences. Our work is able to extract more fine-grained and interpretable edits by leveraging span alignment. Several prior work utilizes the intention to categorize edits and as a clue to understanding the purpose of the revision. Some of their intention taxonomies focus on a specific domain, such as Wikipedia \cite{yang-etal-2017-identifying-semantic, anthonio-etal-2020-wikihowtoimprove} and argumentative essay \cite{zhang-etal-2017-corpus, kashefi2022argrewrite}. The intention taxonomy in our work is built on top of prior literature, with an adaptation to the scientific writing domain.

\section{Conclusion}

In this paper, we present a comprehensive study that investigates the human revision process  in the scientific writing domain. We first introduce {\arxivedits}, a new annotated corpus of 751 full arXiv papers with gold sentence alignment across their multiple versions of revisions, and fine-grained span-level edits together with their underlying intents for 1,000 sentence pairs. Based on this high-quality annotated corpus, we perform a series of data-driven   studies to analyze the common strategies used by the researchers to improve the writing of their papers. In addition, we develop automatic methods to  analyze revision at document-, sentence-, and word-levels. Our annotated dataset, analysis, and automatic system together provide a complete solution for studying text revision in   the scientific writing domain.

\section*{Limitations}
Due to the user groups of  arXiv, our corpus mainly covers research papers in the field of science and engineering, while doesn't contain articles from other areas, such as philosophy and arts. In addition, future work could investigate research papers that are written in  non-English languages.

\section*{Acknowledgement}

We thank three anonymous reviewers for their helpful feedback on this work. We also thank Andrew Duffy and Manish Jois
 for their help with annotations and human evaluation. This research is supported in part by the NSF awards IIS-2144493 and IIS-2055699. The views and conclusions contained herein are those of the authors and should not be interpreted as necessarily representing the official policies, either expressed or implied, of NSF or the U.S. Government. The U.S. Government is authorized to reproduce and distribute reprints for governmental purposes notwithstanding any copyright annotation therein.

\bibliography{anthology,custom}
\bibliographystyle{acl_natbib}

\clearpage
\appendix

\section{Details of Preprocessing}
\label{sec:preprocessing}

We randomly sample 1,000 paper IDs from arXiv that have multiple versions available and download their LaTeX source code for all the versions. During the pre-processing process, we aim to keep each article group complete.  About 105 versions of papers don’t have source code available. After removing them, 959  groups are complete. There are 4 groups of papers using the \texttt{harvmac} package, which will introduce detex errors; after removing them,  955 groups are left. We then remove 162 groups of math-heavy papers and 42 groups of extremely short papers. After the filtering process, 751 complete groups of papers are left. 

Among 335k sentences from all versions in the 751 article groups, we skip aligning  5.8\% sentences  that contain too few words or too many special tokens.
Most of the 5.8\% skipped content is (a) unusually short (<=3 tokens) and math-heavy text (>=60\% special tokens) in math papers or (b) occasional de-tex errors (>1000 char). They don't contain much natural language content that can be analyzed or leveraged. In addition, annotators reported that such text increases the difficulty of aligning sentences.  The criteria are detailed below.

We will skip aligning a paragraph if it meets one of the following criteria:

\begin{itemize}
\vspace{-6pt}
\setlength\itemsep{-3pt}
    \item Contains $<$ 10 tokens.

    \item Contains $>$ 30\% special tokens (inline/block math, citation, and references). 
\end{itemize}

We will skip aligning a sentence if it meets one of the following criteria:
\vspace{-6pt}
\begin{itemize}
\setlength\itemsep{-3pt}
    \item Contains $>$ than 1,000 characters.
    \item Contains $\leq$ than 3 tokens. 
    \item Contains $>$ 60\% special tokens (inline/block math, citation, and references). 
    \item Contains $\geq$ 70\% English characters. 
    \item Ends with ``,'' or ``:''. 
\end{itemize}

We also detect citations, references, inline math symbols, block equations, and present them as special markers which are easier to read for the annotators.  

\section{Details of Paragraph Alignment and the Annotation Process}
\label{Appendix:para_align}
 We design a light-weighted automatic paragraph alignment algorithm based on Jaccard similarity, which can cover 92.1\% of non-identical sentence alignment in the pilot study. The algorithm is shown in Algorithm \ref{alg:paragraph_similarity}. The lengths of the two documents are represented by $k$ and $l$. $d$ denotes the difference of relative position for two paragraphs with indices of $i$ and $j$, where $d(i,j)=|\frac{i}{k}-\frac{j}{l}|$.  The hyperparameters $\tau_1 =  0.28$, $\tau_2 = 0.15$ , $\tau_3 = 0.85$, $\tau_4 = 0.2$ are tuned on the devset. 
 
 At a high level, given an article pair, we first calculate the pairwise similarities for all possible paragraph pairs using  the first block of the algorithm. Paragraph pairs are aligned by the second and third blocks of the algorithm if  their similarity and relative distance reach certain thresholds.

To improve the money efficiency when annotating sentence alignment, we design a hybrid method to collect alignment labels for each sentence pair in the aligned paragraph pairs.   We  found that sentence pairs with Jaccard similarity $> 0.7$ and $<0.2$ can be reliably automatically labeled as \textit{aligned} and \textit{not-aligned}.  The thresholds are determined based on a pilot study and can achieve nearly 100\% precision. Therefore, we automatically labeled the sentence pairs with Jaccard similarity $> 0.7$ and $<0.2$ , and collected human annotation for the rest of the candidate sentence pairs on  the Figure Eight platform. 

\normalem
\begin{algorithm}[pht!]
\small
\SetKwInput{kwInit}{Initialize}

\SetAlgoLined
\kwInit{$alignP$ $\in$ $\mathbb{I}^{k \times l}$ to $0^{k \times l}$}
\kwInit{$simP$ $\in$ $\mathbb{R}^{2 \times k \times l}$ to $0^{2 \times k \times l}$}
\For{$i\leftarrow 1$ \KwTo $k$}
{
    \For{$j\leftarrow 1$ \KwTo $l$}
    {
        \resizebox{1.0\hsize}{!}{
            $ simP[1, i, j] =   \avg\limits_{s_p \in S_i}\Big( \max\limits_{c_q \in C_j} simSent (s_p, c_q) \Big)$ \nonumber
        }

        \resizebox{1.0\hsize}{!}{
            $ simP[2, i, j] =  \avg\limits_{c_p \in C_i}\Big( \max\limits_{s_q \in S_j} simSent (s_p, c_q) \Big)$ \nonumber
        }
    }

}

\SetKwInOut{Input}{Input}
\SetKwInput{kwInit}{Initialize}

\SetAlgoLined

\For{$j\leftarrow 1$ \KwTo $l$}
{
    $ i_{max} = \argmax\limits_{i}  simP[2, i,j] $ \\
    \uIf{$simP[1, i_{max},j] > \tau_1$ and $d(i_{max}, j) < \tau_2$ }{$alignP[i_{max}, j]=1$}
    \uElseIf{$simP[1, i_{max},j] > \tau_3$}
    {$alignP[i_{max}, j]=1$}

}

\For{$i\leftarrow 1$ \KwTo $k$}
{
    $ j_{max} = \argmax\limits_{j}  simP[1, i,j] $ \\
    \uIf{$simP[2, i,j_{max}] > \tau_1$ and $d(i, j_{max}) < \tau_4$ }{$alignP[i, j_{max}]=1$}
    \uElseIf{$simP[2, i,j_{max}] > \tau_3$}
    {$alignP[i, j_{max}]=1$}

}

\Return  $alignP$

\caption{Paragraph Alignment Algorithm}

\label{alg:paragraph_similarity}
\end{algorithm}

\section{Experiment Details}
Our experiments are run on 4$\times$A40 GPUs. The implementation and hyperparameter tuning process are detailed below, where the one marked with $*$ performs best. We perform 3 runs for each setting, and average the performance. We use scikit-learn package to calculate the precision, recall and F1.\footnote{\url{https://scikit-learn.org/stable/modules/generated/sklearn.metrics.classification_report.html}}

\paragraph{Sentence Alignment.} We use the author's implementation of the  neural CRF sentence alignment model and initialize it  with the pretrained SciBERT-based-uncased encoder \cite{beltagy-etal-2019-scibert}. We tune the learning rate in \{1e-5, 3e-5$^*$, 5e-5\} based on F1 on the devset. The model is trained within 1.5 hours.

\paragraph{Intention Classification.} We use the Huggingface\footnote{\url{https://huggingface.co/}} implementation of the T5 model, and use the author's implementation of the PURE model. We initialized the PURE model with SciBERT-based-cased encoder \cite{beltagy-etal-2019-scibert}. We tune the learning rate in \{1e-5, 3e-5, 5e-5, 7e-5$^*$\} based on F1 on the devset. Both models are trained within 1 hour.

\paragraph{Edits Extraction.} We use the original author's implementations for the neural semi-CRF word alignment model and the QA-Align model. We initialize the semi-CRF model with SpanBERT-large encoder \cite{joshi-etal-2020-spanbert} and initialize the QA-Align model with SciBERT-based-uncased encoder \cite{beltagy-etal-2019-scibert}. We use the default hyperparameters for both models. The semi-CRF model takes about 10 hours to train, and the QA-Align model takes about 3 hours to train.

\clearpage
\section{Crowdsourcing Annotation Interface}
\label{appendix:human-annotation}
\subsection{Screenshot of the Instructions}

\noindent\begin{minipage}{\textwidth}
    \centering

    \includegraphics[width=0.95\textwidth]{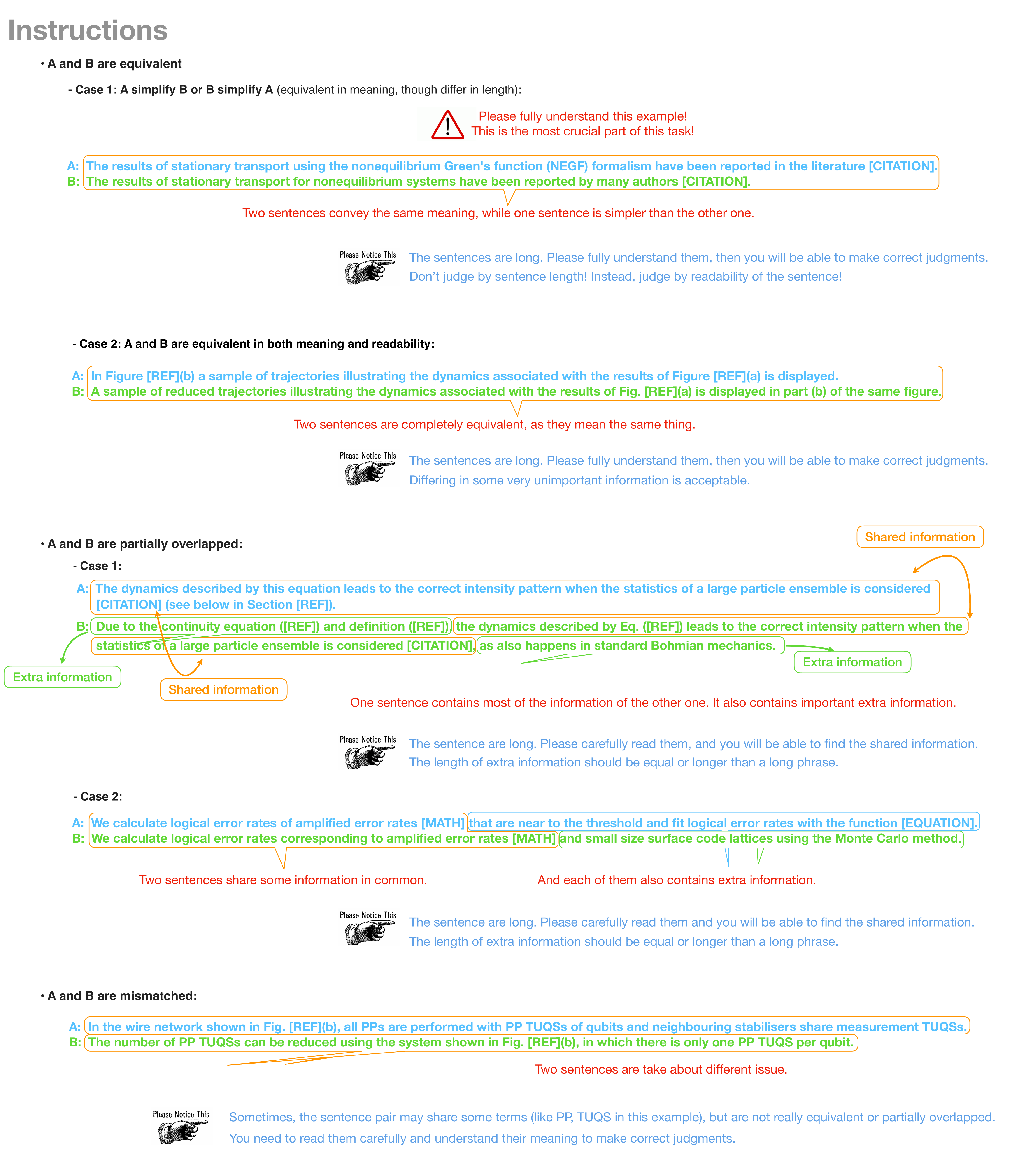}
    \captionof{figure}{Instructions for our crowdsourcing annotation of sentence alignments on the Figure Eight platform.}
    \label{fig:annotation-}
\end{minipage}

\clearpage
\subsection{Screenshot of the Interface}

\noindent\begin{minipage}{\textwidth}
    \centering

    \includegraphics[width=0.8\textwidth]{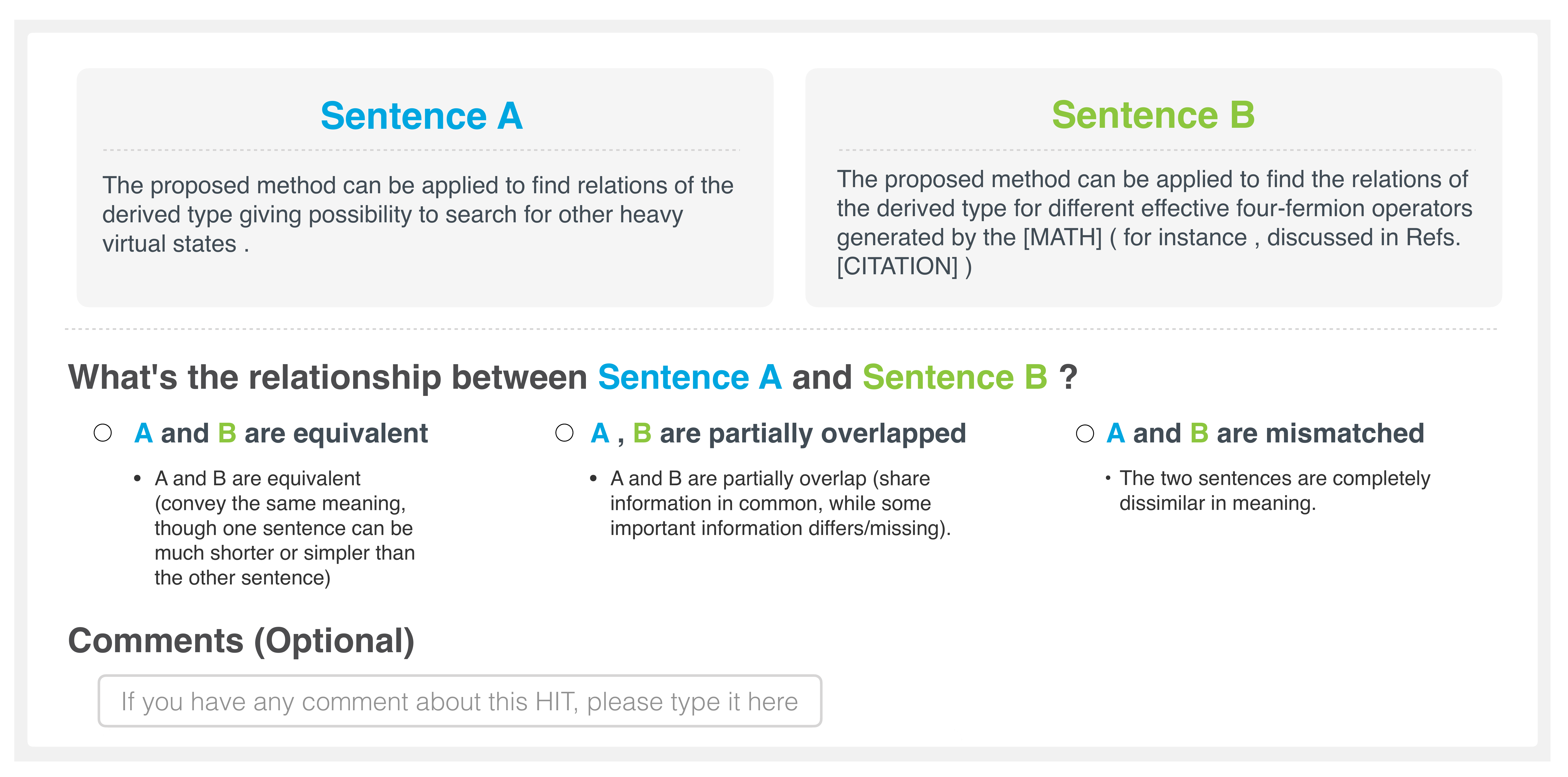}
    \captionof{figure}{Interface for our crowdsourcing annotation of sentence alignments on the Figure Eight platform.} 
    \label{fig:interface-}
\end{minipage}

\end{document}